\documentclass[
twocolumn,
hf,
]{ceurart}

\usepackage{pgfplots}
\pgfplotsset{compat=1.17} 

\usepackage{listings}
\begin{document}

\copyrightyear{2024}
\copyrightclause{Copyright for this paper by its authors.
  Use permitted under Creative Commons License Attribution 4.0
  International (CC BY 4.0).}

\conference{AAAI-25 Defactify 4 Workshop,
 Feb 2025,  Philadelphia, Pennsylvania, USA}

\title{Team NYCU at Defactify4: Robust Detection and Source Identification of AI-Generated Images Using CNN and CLIP-Based Models}


\author[1]{Tsan-Tsung, Yang}[%
email=alexyang0826@hotmail.com,
]
\cormark[1]
\fnmark[1]
\address[1]{Department of Computer Science, National Yang Ming Chiao Tung University, Hsinchu, Taiwan}

\author[2]{I-Wei, Chen}[%
email=ken12300326@gmail.com,
]
\address[2]{Department of Electronics and Electrical Engineering , National Yang Ming Chiao Tung University, Hsinchu, Taiwan}
\fnmark[1]

\author[1]{Kuan-Ting, Chen}[%
email=larrybrown901120@gmail.com,
]
\fnmark[1]

\author[1]{Shang-Hsuan, Chiang}[%
email=andy10801@gmail.com,
]
\fnmark[2]

\author[1]{Wen-Chih, Peng}[%
email=wcpeng@cs.nycu.edu.tw,
]
\fnmark[2]

\cortext[1]{Corresponding author.}
\fntext[1]{These authors contributed equally.}

\begin{abstract}
  With the rapid advancement of generative AI, AI-generated images have become increasingly realistic, raising concerns about creativity, misinformation, and content authenticity. Detecting such images and identifying their source models has become a critical challenge in ensuring the integrity of digital media. This paper tackles the detection of AI-generated images and identifying their source models using CNN and CLIP-ViT classifiers. For the CNN-based classifier, we leverage EfficientNet-B0 as the backbone and feed with RGB channels, frequency features, and reconstruction errors, while for CLIP-ViT, we adopt a pretrained CLIP image encoder to extract image features and SVM to perform classification. Evaluated on the Defactify 4 dataset, our methods demonstrate strong performance in both tasks, with CLIP-ViT showing superior robustness to image perturbations. Compared to baselines like AEROBLADE and OCC-CLIP, our approach achieves competitive results. Notably, our method ranked Top-3 overall in the Defactify 4 competition, highlighting its effectiveness and generalizability. All of our implementations can be found in \href{https://github.com/uuugaga/Defactify_4}{https://github.com/uuugaga/Defactify\_4}
\end{abstract}


\maketitle

\section{Introduction}

Recent advancements in text-to-image generation models have made it possible to produce high-quality images from simple prompts. This development poses challenges to content creators and raises concerns about the authenticity of online content. Over the years, various generative models, including Generative Adversarial Networks (GANs) [1], Variational Autoencoders (VAEs) [2], Stable Diffusion [3], DALL·E [4], and Midjourney [5], have emerged and continuously improved.

To effectively understand and regulate AI-generated images, it is crucial not only to detect whether the content is real or fake but also to identify the specific model used to generate it. However, research efforts on the identification of the source model remain limited. The Defactify 4 workshop dataset [6], which includes state-of-the-art text-to-image models such as Stable Diffusion, DALL·E, and Midjourney, addresses this gap by offering a benchmark for two tasks: (A) classifying AI-generated content and (B) identifying the source models. The dataset also accounts for real-world scenarios where images are often modified differently. To emphasize the generalizability of the detection methods, this dataset also put some perturbations on the generated images.

In this paper, we adopt two primary approaches—CNN-based and CLIP-based classifiers—to evaluate their performance on both tasks. Additionally, we conduct comprehensive experiments, including baseline comparisons, robustness evaluations under perturbations, and ablation studies on data augmentation. Our findings can be summarized as follows:

\begin{itemize}
\item Both CLIP-ViT and CNN-based methods effectively detect AI-generated content and identify the source model. However, CLIP-ViT demonstrates superior robustness when images are subjected to perturbations in real-world scenarios.
\item Our methods achieve competitive performance or even better compared to strong baselines, including AEROBLADE [1] and OCC-CLIP [2]. 
\item Ablation studies reveal that applying perturbations—such as Gaussian noise, JPEG compression, brightness reduction, and Gaussian blurring—during training significantly improves the models' generalization ability. 
\end{itemize}

These results underscore the importance of robust model design and data augmentation for detecting and identifying AI-generated images in practical applications.

\begin{figure*}
    \centering
    \includegraphics[width=1\linewidth]{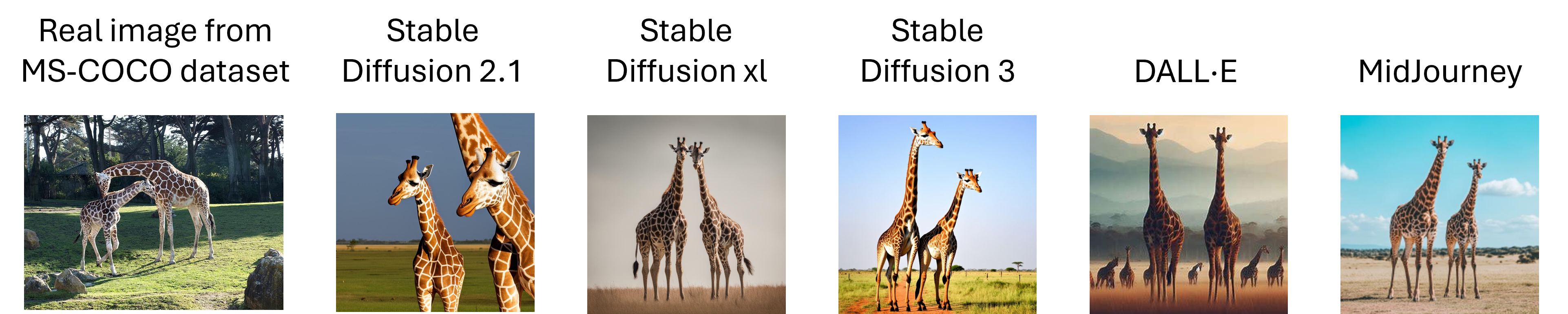}
    \caption{Given the same prompt, "Two tall giraffes standing next to each other on a field", the different models can generate different image content and still keep the text consistency. The left one is the real image, while the others are all generated by text-to-image models. It is clear that each model has its own "assumption" and "style" on the given prompt, which can be captured as the model's features.}
    \label{fig:data_samples}
\end{figure*}

\section{Related Works}

\subsection{Revolution on Image Generation}
The landscape of image generation has undergone a dramatic transformation with the rise of deep learning techniques. Initially, Generative Adversarial Networks (GANs) \cite{gan} revolutionized the field by introducing a two-network architecture—a generator and a discriminator—that enables the production of realistic images through adversarial training. Another milestone in this evolution was the development of Variational Autoencoders (VAEs) \cite{auto_encoder}, which employed a probabilistic approach to generative modeling. VAEs utilized a likelihood-based objective to learn a lower-dimensional latent space representation of input data.

More recently, the focus has shifted towards diffusion models, which have demonstrated significant improvements in image quality and authenticity. Unlike GANs and VAEs, which rely on adversarial learning and latent space encoding, respectively, Denoising Diffusion Probabilistic Models (DDPMs) \cite{ddpm} aim to learn a denoising process. Starting with pure noise, DDPMs iteratively refine it through a series of denoising steps, transforming random noise into coherent, meaningful images. Remarkably, DDPMs can generate images with greater diversity and fewer artifacts than GANs and VAEs, often producing more realistic and visually appealing results.

Building on these advancements, the field has also seen a rise in text-to-image generation, with models such as DALL·E \cite{dalle}, Imagen \cite{imagen}, Stable Diffusion \cite{stable_diffusion}, Midjourney \cite{midjourney} enabling users to generate detailed images from simple textual prompts. These models leverage pre-trained text encoders and VAEs to map both text and images into a shared latent space, where they then perform a diffusion process on the latent representations rather than on the original images. These models can now generate images that are visually compelling and contextually relevant to the prompts, often producing results that are remarkably close to the user's original vision.

\subsection{Classifying AI-Generated Images}
Image generation is a double-edged sword: it offers excellent potential across various fields, from creative industries to healthcare research, yet it also facilitates the propagation of misinformation on social media. This underscores the importance of detecting AI-generated content to preserve the integrity of information. 

One intuitive approach is to learn a binary classifier, which directly detects whether the image is real or fake. Marra et al. \cite{CNN_detector} investigate various convolutional neural network (CNN)-based models, for identifying GAN-generated content. Their findings suggest that CNN-based models effectively detect images generated by GANs. Cozzolino et al. \cite{clip_svm} applied CLIP \cite{clip} to encode image captions as the fake content and train the SVM classifier to detect the AI-generated image. They demonstrated that CLIP features provide excellent generalization, achieving strong performance even with a limited number of examples. Instead of directly training, Alam et al. \cite{fft} aimed to add more extra feature to the classifier. They converted RGB channels into YCbCr channels and applied Spatial Fourier Transformation to capture spatial shifts. By combining deep neural networks with feature fusion, their method outperforms existing state-of-the-art techniques.

Other researchers assume that AI-generated content can be easily reconstructed by AI-modules. Wang et al. \cite{dire} found that DM images can be approximately reconstructed by the diffusion model, but real images cannot. The difference between the reconstructed image and the original image is recorded as Diffusion Reconstruction Error (DIRE). Then DIRE can be used as a feature for training to determine whether it is real or fake, and the generalization will be much higher. Ricker et al. \cite{aeroblade} propose the AEROBLADE framework, performing reconstruction on auto-encoders. They use VGG network's hidden layer to measure the reconstruction distance, which is called $\text{LPIPS}_{2}$ distance. Their experiment result is awesome without any training process.

\subsection{Source Model Identification}
To effectively understand, regulate, and categorize AI-generated content, it is essential to identify the model used to generate the image accurately. However, current research primarily focuses on distinguishing between real and AI-generated images. For source model identification, some approaches require direct access to or modifications of the source model. This is not practical in real-world scenarios since some generative models are not publicly accessible, like MidJourney and DALL·E. The most recent work on source model identification, OCC-CLIP \cite{occ_clip}, modifies the problem into a few-shot learning setting, achieving good performance and scalability to larger datasets. By combining prompt learning with adversarial augmentation, OCC-CLIP keeps the original CLIP \cite{clip} parameters fixed and tunes only a learnable context. Additionally, it employs multiple one-class classifiers to predict multi-class targets, yielding excellent results in both binary and multi-class classification tasks.

\section{Method}

\subsection{Problem Formulation}
Let $D = \{(x_i, y_i)\}_{i=1}^{N}$ denote the image-label pairs in the dataset, where the $i$-th sample consists of an image $x_i$ and a corresponding label $y_i$. For task A, the goal is to detect whether the image is AI-generated or not. In this case, $y_i$ is a binary label, where $y_i \in \{\text{real}, \text{fake}\}$. In contrast, task B is more challenging, as it requires identifying the source model of the image. Therefore, $y_i$ now represents the source model label; that is, $y_i \in \{\text{real}, \text{SD}_{2.1}, \text{SD}_{xl}, \text{SD}_3, \text{DALL·E}, \text{MidJourney}\}$, where SD is an abbreviation of Stable Diffusion and the subscript is the model's version.

Formally, the goal is to learn a decision function $f_\theta(x)$ that minimizes the classification error. It can be formulated as:

\begin{equation} 
\theta^* = \arg\min_{\theta} \ \mathbb{E} \left[ \mathcal{L}(f_{\theta}(x), y) \right]
\label{eq:objective}
\end{equation}
where $\mathcal{L}$ is a error measurement of the prediction and ground truth and $\theta$ is the model parameters.

\subsection{CNN or CLIP-ViT}
To explore the effectiveness of CNN-based and CLIP-based methods on AI-generated image detection, we select EfficientNet-B0 \cite{efficientnet} and CLIP-ViT \cite{clip} to compare their performance.

\subsubsection{EfficientNet-B0}
EfficientNet-B0 \cite{efficientnet} is a convolutional neural network (CNN) that balances accuracy and efficiency by using a compound scaling method. This approach uniformly scales the network's depth, width, and resolution, enabling high performance with relatively fewer parameters compared to traditional CNNs. 

To enrich the model with more detailed information, we adopt feature augmentation strategies inspired by Alam et al. \cite{fft} and Ricker et al. \cite{aeroblade}. Specifically, we incorporate additional input features—frequency information and reconstruction error—alongside the standard RGB image. Formally, we can denote the original RGB image as $I$, the reconstruction error as $E$ and the frequency domain feature as $F$, where $I \in \mathbb{R}^{h\times w \times3}$, $E, F \in \mathbb{R}^{h\times w \times1}$ and $h, w$ represent the image's height and width. By concatenating these features along the channel dimension, the final input to the model becomes:
\begin{equation}
X = \text{concat}(I,E,F) \in \mathbb{R}^{h \times w \times 5}
\end{equation}
where $\text{concat}(\cdot)$ denotes channel-wise concatenation.

For EfficientNet-B0 method, we adopt the cross-entropy loss to find the approximately solution of $\theta^{*}$, the original classification error $\mathcal{L}$ from equation~\ref{eq:objective} can be rewritten as:
\begin{equation}
\mathcal{L}(\theta)=-\sum_{i=1}^{N}\sum_{c=1}^{C}\mathbb{I}(y_i=c)\log p(f_\theta(X_i)=c|X_i,\theta)
\end{equation}
where $X_i$ is the augmented input for the i-th sample, $\mathbb{I}(\cdot)$ is the indicator function and $f_\theta(X_i)$ is the model output given the input $X_i$.

\subsubsection{CLIP-ViT}
CLIP-ViT \cite{clip} combines the power of Vision Transformers (ViT) with a contrastive pretraining framework. Designed for multi-modal learning, CLIP aligns images and text in a shared latent space, enabling robust zero-shot and transfer learning capabilities.

In our approach, we utilize the pretrained CLIP-ViT model without prompt for image feature extraction. Given an input image $x_i$, the CLIP encoder $f_\text{CLIP}$ maps RGB images into a high-dimensional feature space with feature size $d$:
\begin{equation}
z_i=f_{\text{CLIP}}(X_i) \in \mathbb{R}^{d}
\end{equation}

For classification, we apply a Support Vector Machine (SVM) \cite{svm} with a Radial Basis Function (RBF) kernel to separate different classes. The decision function of the SVM is defined as:
\begin{equation}
f_{\theta}(z)=\text{sign} \left ( \sum_{i=1}^{N}\alpha_i y_i K \left ( z, z_i \right ) +b \right )
\end{equation}
where $\alpha_i$ are the Lagrange multipliers, $K(z, z_i)=\exp \left ( -\gamma \cdot \lVert z-z_i \rVert ^2 \right )$ is the RBF kernel, $\gamma$ controls the kernel's spread, $b$ is the bias term.

The SVM optimization problem aims to maximize the margin while allowing soft-margin classification, the equation~\ref{eq:objective} can be written as:
\begin{equation}
\mathcal{L}(\theta)=\frac{1}{2}\lVert \theta \rVert ^2 + C\sum_{i=1}^{N}\xi_i
\end{equation}
subject to:
\begin{equation}
y_i\cdot f_\theta(z_i) \geq 1 - \xi_i, \;\; \xi_i \geq 0
\end{equation}
where $C$ balances margin maximization and classification error and $\xi_i$ are slack variables for misclassification tolerance. For multi-class classification, we employ the One-vs-Rest (OvR) strategy, where a separate binary SVM classifier is trained for each class $c$ against all other classes

\section{Experiments}

\subsection{Setup}
\subsubsection{Data} 
We utilize the dataset provided in this shared task \cite{roy-2025-defactify-dataset-image} \cite{roy-2025-defactify-overview-image} to evaluate the performance of our method. The dataset consists of four parts: training, validation, testing, and final testing. The first three parts are standard splits of the dataset, while the final testing part introduces perturbations to the original images. Detailed statistics for all four parts can be found in Table~\ref{tab:dataset}, and the training set is a balanced dataset, with each label containing exactly 7000 samples. Fig~\ref{fig:data_samples} gives an example of this dataset.

To improve our model's robustness, we apply the following data augmentation techniques to the original training dataset:
\begin{itemize}
\item \textbf{Compression}: Considering real-world scenarios where images are often compressed in JPEG format, we apply JPEG compression with a quality parameter of 50.
\item \textbf{Blurring}: We apply Gaussian blurring with a sigma value of 5 and a kernel size of (5, 5).
\item \textbf{Noise Perturbation}: We add small Gaussian noise to the original image with a standard deviation of 0.3.
\item \textbf{Brightness Transformation}: We adjust the image brightness using a brightness factor of 0.5.
\end{itemize}

\begin{table}[ht]
  \caption{Statistics of the shared task dataset.}
  \label{tab:dataset}
  \begin{tabular}{c|c}
    \toprule
    Split Name & Number of Samples \\
    \midrule
    Training & 42000 \\
    Validation & 9000 \\
    Testing & 9000 \\
    Final Testing & 45000 \\
    \bottomrule
  \end{tabular}
\end{table}

\subsubsection{Implementation Details}

For the EfficientNet method, all RGB images are resized to $(512, 512, 3)$ and further processed to extract additional features: frequency and reconstruction error. The frequency feature is obtained by converting the image to grayscale, applying a 2D Fast Fourier Transform (FFT), shifting the zero-frequency component to the center, and computing the logarithmic magnitude spectrum. To calculate the reconstruction error, we use the VAE from the \texttt{runwayml/stable-diffusion-v1-5} model to reconstruct each image and compute the pixel-wise absolute difference between the reconstructed and original images. These two features are concatenated with the original RGB channels, resulting in a five-channel input of size $(512, 512, 5)$. The classifier is trained from scratch using the EfficientNet-B0 backbone with the Adam optimizer, a learning rate of $1 \times 10^{-4}$, and a total of 30 epochs.

For CLIP-ViT, we select the pre-trained model \texttt{openai/clip-vit-base-patch16} to extract the image features. After that, we trained a SVM classifier based on the image features. We performed a grid search over the following hyperparameters: $c \in \{ 0.001,0.01,0.1,1,10,100,1000 \}$ and $\gamma \in \{ 0.001,0.01,0.1,1,10,100,1000 \}$. The final predictions are based on the best-performing hyperparameter set identified through grid search.

Our experiments were conducted on a machine equipped with 48 AMD Ryzen Threadripper 3960X 24-Core Processors, 237GB of RAM and 2 NVIDIA GeForce RTX 3090 GPUs. The source code is publicly available at: \href{https://github.com/uuugaga/Defactify_4}{https://github.com/uuugaga/Defactify\_4}

\subsection{Evaluation}
In these two tasks, we select both accuracy and macro-f1 score as our metrics in both tasks. Since task A is a little bit imbalanced on real images, macro-f1 score is much more reliable on these two tasks.

\subsubsection{Comparison to Baselines}
We compare our method with some state-of-the-art baselines by modifying the source code released by the original authors. For both tasks, we select AEROBLADE \cite{aeroblade} and OCC-CLIP \cite{occ_clip} as our baseline methods due to their simplicity and effectiveness. Despite their relatively straightforward approaches, these methods have shown strong performance in real-world scenarios. 

\begin{figure}
    \centering
    \includegraphics[width=1\linewidth]{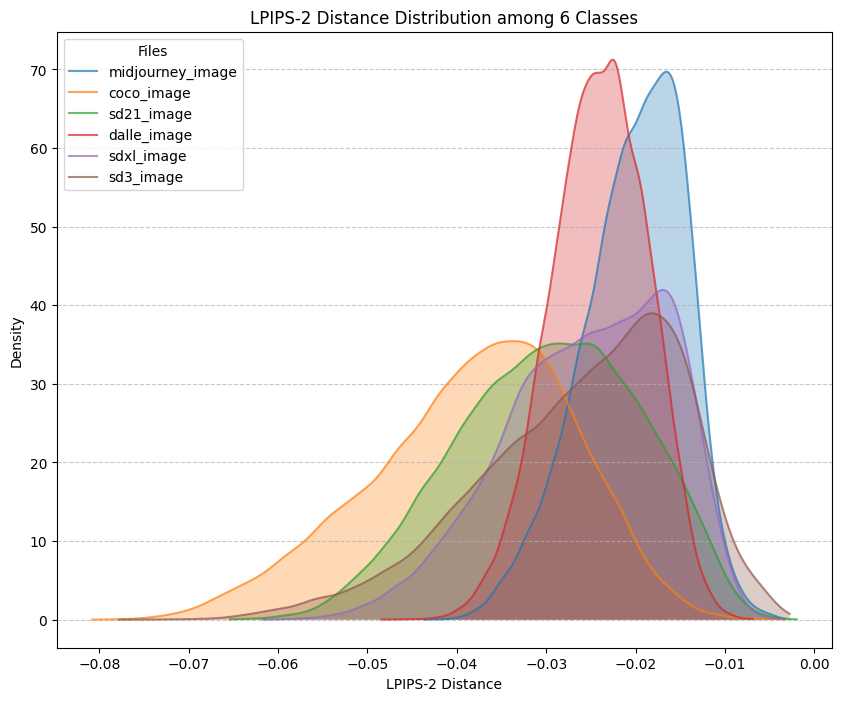}
    \caption{The $\text{LPIPS}_{2}$ distance distribution of generated content from different models. It's obvious that the distribution has a lot of overlapped areas, which, in turn, indicates that AEROBLADE's performance would not be great.}
    \label{fig:aeroblade_histogram}
\end{figure}

AEROBLADE \cite{aeroblade} applies three types of pre-trained autoencoders for reconstruction. It assumes that the reconstruction distance for AI-generated content differs from that of real images. For our implementation of AEROBLADE, we follow the parameter settings from the original authors, with modifications to the batch size (set to 16) and the image resolution (set to 512 × 512). Additionally, we adopt the original experimental results and choose the $\text{LPIPS}_{2}$ distance as the metric for reconstruction distance. Using the training set, we plot a kernel density estimation of the $\text{LPIPS}_{2}$ distance, as shown in Fig.~\ref{fig:aeroblade_histogram}. We set the distance threshold to -0.035 to classify the images as either real or fake. Since AEROBLADE in the original paper is only suitable for binary classification, we only compare with it in task A.

OCC-CLIP \cite{occ_clip} is another baseline. It freezes the CLIP model's weight by only tuning the learnable context. It combines both prompt learning and adversarial augmentation in a few-shot source model identification. We alter the context length to 32 and follow the original setting as a multiple two-class classifiers. In this case, We train 5 classifiers to detect whether it is real or one of the target models, and each of them is trained on 7000 real images and 7000 target fake images. The final prediction is followed by the authors of OCC-CLIP:

\begin{equation}
\hat{y}_{i}=\left\{
\begin{aligned}
j & , & \max_{j\in1,...,5} p(\hat{y}_i=1; x_i, \theta_{j}) > 0.5\\
0 & , & otherwise.
\end{aligned}
\right.
\end{equation}
where $\theta_j$ is the $j$-th OCC-CLIP for classifying the content is real or generated from model $j$. It means that the final prediction is made by selecting the target model with the highest probability, or classifying it as real if none of the classifiers exceed the threshold 0.5.

Table~\ref{tab:baseline} shows that our method outperforms AEROBLADE and keeps almost the same performance as OCC-CLIP in task A. However, in task B, our method surpasses OCC-CLIP by approximately 0.12 in accuracy and macro-f1, which corresponds to a \textbf{14\%} improvement.

\begin{table}[ht]
  \caption{Compare our methods with AEROBLADE and OCC-CLIP in tasks A and B on the validation set, with accuracy and macro-f1 score. The bold font highlights the best performance, while the underlined font indicates the second-best result. Both of our methods achieve competitive results on Task A and outperform all other baseline methods on Task B.}
  \label{tab:baseline}
  \begin{tabular}{c|c|c|c|c}
    \toprule
    \multirow{2}{*}{\parbox[c]{2cm}{\centering Method}} & \multicolumn{2}{c}{Task A} & \multicolumn{2}{c}{Task B} \\ 
           & Acc. & F1 & Acc. & F1 \\
    \midrule
    AEROBLADE & 0.8149 & 0.6986 & - & - \\
    OCC-CLIP  & \textbf{0.9934} & \textbf{0.9881} & 0.8693 & 0.8721 \\
    \midrule
    Ours: EfficientNet-B0  & \underline{0.9849} & \underline{0.9833} & \textbf{0.9951} & \textbf{0.9951} \\
    Ours: CLIP-ViT & 0.9421 & 0.9421 & \underline{0.9377} & \underline{0.9317} \\
    \bottomrule
  \end{tabular}
\end{table}

\subsubsection{Robustness to Perturbations}

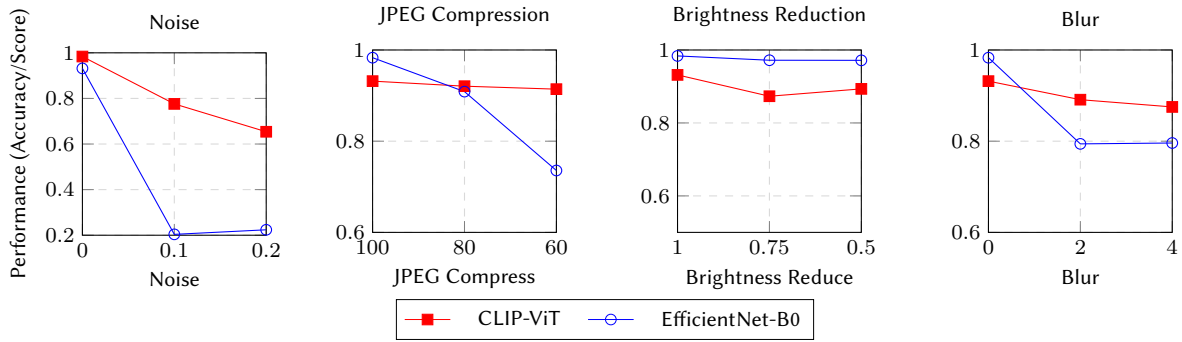
\begin{figure*}[ht!]
    \centering
    \begin{minipage}{0.24\textwidth}
        \centering
        \begin{tikzpicture}
            \begin{axis}[
                title={Noise},
                xlabel={Noise},
                ylabel={Performance (Accuracy/Score)},
                xmin=0, xmax=0.2,
                ymin=0.2, ymax=1.0,
                xtick={0, 0.1, 0.2},
                ytick={0.2, 0.4, 0.6, 0.8, 1.0},
                grid=major,
                grid style={dashed, gray!30},
                width=4cm, 
                height=4cm, 
            ]
                \addplot[color=red, mark=square*] coordinates {
                    (0, 0.9837)
                    (0.1, 0.7760)
                    (0.2, 0.6538)
                };

                \addplot[color=blue, mark=o] coordinates {
                    (0, 0.9317)
                    (0.1, 0.2038)
                    (0.2, 0.2238)
                };
                
            \end{axis}
        \end{tikzpicture}
    \end{minipage}
    \hfill
    \begin{minipage}{0.24\textwidth}
        \centering
        \begin{tikzpicture}
            \begin{axis}[
                title={JPEG Compression},
                xlabel={JPEG Compress},
                xmin=60, xmax=100,
                ymin=0.6, ymax=1.0,
                xtick={60, 80, 100},
                ytick={0.6, 0.8, 1.0},
                x dir=reverse,
                grid=major,
                grid style={dashed, gray!30},
                width=4cm, 
                height=4cm, 
            ]
                \addplot[color=red, mark=square*] coordinates {
                    (100, 0.9317)
                    (80, 0.9206)
                    (60, 0.9141)
                };

                \addplot[color=blue, mark=o] coordinates {
                    (100, 0.9831)
                    (80, 0.9087)
                    (60, 0.7357)
                };
            \end{axis}
        \end{tikzpicture}
    \end{minipage}
    \hfill
    \begin{minipage}{0.24\textwidth}
        \centering
        \begin{tikzpicture}
            \begin{axis}[
                title={Brightness Reduction},
                xlabel={Brightness Reduce},
                xmin=0.5, xmax=1,
                ymin=0.5, ymax=1.0,
                xtick={0.5, 0.75, 1},
                ytick={0.6, 0.8, 1.0},
                x dir=reverse,
                grid=major,
                grid style={dashed, gray!30},
                width=4cm, 
                height=4cm, 
            ]
                \addplot[color=red, mark=square*] coordinates {
                    (1, 0.9317)
                    (0.75, 0.8732)
                    (0.5, 0.8934)
                };

                \addplot[color=blue, mark=o] coordinates {
                    (1, 0.9837)
                    (0.75, 0.9719)
                    (0.5, 0.9716)
                };
            \end{axis}
        \end{tikzpicture}
    \end{minipage}
    \hfill
    \begin{minipage}{0.24\textwidth}
        \centering
        \begin{tikzpicture}
            \begin{axis}[
                title={Blur},
                xlabel={Blur},
                xmin=0, xmax=4,
                ymin=0.6, ymax=1.0,
                xtick={0, 2, 4},
                ytick={0.6, 0.8, 1.0},
                xmode=linear,
                grid=major,
                grid style={dashed, gray!30},
                width=4cm, 
                height=4cm, 
            ]
                \addplot[color=red, mark=square*] coordinates {
                    (0, 0.9317)
                    (2, 0.8910)
                    (4, 0.8751)
                };

                \addplot[color=blue, mark=o] coordinates {
                    (0, 0.9831)
                    (2, 0.7941)
                    (4, 0.7960)
                };
            \end{axis}
        \end{tikzpicture}
    \end{minipage}
    \vspace{0.1cm} 
    \
    \begin{tikzpicture}
        \begin{axis}[
            hide axis,
            xmin=0, xmax=1,
            ymin=0, ymax=1,
            legend columns=2,
            legend style={
                at={(0.5, -0.05)}, 
                anchor=north,
                draw=black,         
                line width=0.25pt,
                font=\small,
                column sep=1em
            }
        ]
            \addlegendimage{color=red, mark=square*}
            \addlegendentry{CLIP-ViT}
            
            \addlegendimage{color=blue, mark=o}
            \addlegendentry{EfficientNet-B0}
        \end{axis}
    \end{tikzpicture}
    \vspace{0.2cm}
    \caption{The generalization of CLIP-ViT and EfficientNet on different perturbations. The red square line indicates the CLIP-ViT and the blue circle one indicates EfficientNet. The result shows that CLIP-ViT's performance is better while EfficientNet's performance drops dramatically.}
    \label{fig:generalization}
\end{figure*}
\begin{figure*}[ht!]
    \centering
    \begin{minipage}{0.24\textwidth}
        \centering
        \begin{tikzpicture}
            \begin{axis}[
                title={Noise},
                xlabel={Noise},
                ylabel={Performance (Accuracy/Score)},
                xmin=0, xmax=0.2,
                ymin=0.2, ymax=1.0,
                xtick={0, 0.1, 0.2},
                ytick={0.2, 0.4, 0.6, 0.8, 1.0},
                grid=major,
                grid style={dashed, gray!30},
                width=4cm,
                height=4cm
            ]
                \addplot[color=red, mark=square*] coordinates {
                    (0, 0.9837)
                    (0.1, 0.7760)
                    (0.2, 0.6538)
                };

                \addplot[color=blue, mark=o] coordinates {
                    (0, 0.9421)
                    (0.1, 0.3398)
                    (0.2, 0.2099)
                };
            \end{axis}
        \end{tikzpicture}
    \end{minipage}
    \hfill
    \begin{minipage}{0.24\textwidth}
        \centering
        \begin{tikzpicture}
            \begin{axis}[
                title={JPEG Compression},
                xlabel={JPEG Compress},
                xmin=60, xmax=100,
                ymin=0.6, ymax=1.0,
                xtick={60, 80, 100},
                ytick={0.6, 0.8, 1.0},
                x dir=reverse,
                grid=major,
                grid style={dashed, gray!30},
                width=4cm,
                height=4cm
            ]
                \addplot[color=red, mark=square*] coordinates {
                    (100, 0.9317)
                    (80, 0.9206)
                    (60, 0.9141)
                };

                \addplot[color=blue, mark=o] coordinates {
                    (100, 0.9421)
                    (80, 0.9149)
                    (60, 0.8689)
                };
            \end{axis}
        \end{tikzpicture}

    \end{minipage}
    \hfill
    \begin{minipage}{0.24\textwidth}
        \centering
        \begin{tikzpicture}
            \begin{axis}[
                title={Brightness Reduction},
                xlabel={Brightness Reduce},
                xmin=0.5, xmax=1,
                ymin=0.5, ymax=1.0,
                xtick={0.5, 0.75, 1},
                ytick={0.6, 0.8, 1.0},
                x dir=reverse,
                grid=major,
                grid style={dashed, gray!30},
                width=4cm,
                height=4cm
            ]
                \addplot[color=red, mark=square*] coordinates {
                    (1, 0.9317)
                    (0.75, 0.8732)
                    (0.5, 0.8934)
                };

                \addplot[color=blue, mark=o] coordinates {
                    (1, 0.9421)
                    (0.75, 0.9066)
                    (0.5, 0.8295)
                };
            \end{axis}
        \end{tikzpicture}

    \end{minipage}
    \hfill
    \begin{minipage}{0.24\textwidth}
        \centering
        \begin{tikzpicture}
            \begin{axis}[
                title={Blur},
                xlabel={Blur},
                xmin=0, xmax=4,
                ymin=0.6, ymax=1.0,
                xtick={0, 2, 4},
                ytick={0.6, 0.8, 1.0},
                xmode=linear,
                grid=major,
                grid style={dashed, gray!30},
                width=4cm,
                height=4cm
            ]
                \addplot[color=red, mark=square*] coordinates {
                    (0, 0.9317)
                    (2, 0.8910)
                    (4, 0.8751)
                };

                \addplot[color=blue, mark=o] coordinates {
                    (0, 0.9421)
                    (2, 0.8741)
                    (4, 0.8060)
                };
            \end{axis}
        \end{tikzpicture}
    \end{minipage}
    \vspace{0.1cm} 
    \
    \begin{tikzpicture}
        \begin{axis}[
            hide axis,
            xmin=0, xmax=1,
            ymin=0, ymax=1,
            legend columns=2,
            legend style={
                at={(0.5, -0.05)}, 
                anchor=north,
                draw=black,         
                line width=0.25pt,
                font=\small,
                column sep=1em
            }
        ]
            \addlegendimage{color=red, mark=square*}
            \addlegendentry{W/ Data Augmentation}
            
            \addlegendimage{color=blue, mark=o}
            \addlegendentry{W/o Data Augmentation}
        \end{axis}
    \end{tikzpicture}
    \vspace{0.2cm}
    \caption{The importance of data augmentation. The red square line indicates training with data augmentation and the blue circle one indicates training without data augmentation.}
    \label{fig:data_aug}
\end{figure*}
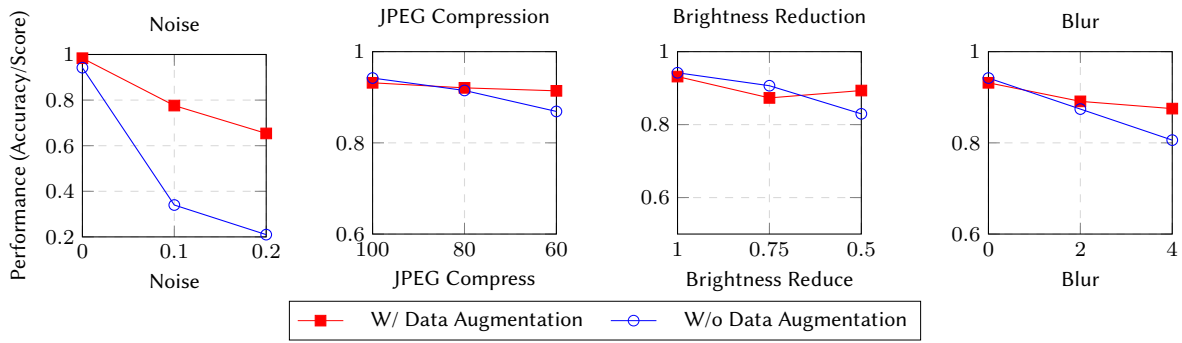

In real-world scenarios, images are often subjected to various processing steps, such as compression, noise addition, brightness adjustments, or blurring, especially during online sharing or editing. The results presented in Figure \ref{fig:generalization} highlight the robustness of models when faced with such perturbations. Across all experiments, the CLIP-ViT backbone model consistently demonstrated superior performance compared to the EfficientNet backbone. For instance, in the noise test (the first sub-figure), CLIP-ViT maintained high accuracy even with increasing noise levels, while EfficientNet suffered a significant performance degradation. Similarly, in JPEG compression (the second sub-figure) and brightness reduction (the third sub-figure), CLIP-ViT achieved more stable accuracy across different perturbation intensities. The blur test (the last sub-figure) further underscores CLIP-ViT's robustness, maintaining competitive performance even under severe blurring conditions. These results emphasize the importance of designing robust models capable of handling real-world image variations effectively.

\subsubsection{Ablation Study: Importance of Data Augmentation}

The results of the ablation study, as illustrated in Figure~\ref{fig:data_aug}, demonstrate the significant impact of data enhancement on model performance under various perturbations. For example, in the noise experiment (the first sub-figure), models with data augmentation consistently outperformed those without, achieving much higher accuracy across all noise levels. Similarly, in the JPEG compression test (the second sub-figure), data-augmented models exhibited superior performance, particularly at higher compression levels. Brightness reduction (the third sub-figure) and blur experiments (Figure the last sub-figure) also highlighted the robustness introduced by data augmentation, with models showing improved accuracy and stability when augmentation techniques were employed. These findings underline the critical role of data augmentation in mitigating performance degradation caused by common perturbations, thereby enhancing the model's generalizability and robustness in real-world scenarios.

\section{Conclusion}

This paper evaluated CNN-based and CLIP-based methods for detecting AI-generated images and identifying their source models using the Defactify 4 workshop dataset. Our key findings include:

\begin{itemize} \item Both EfficientNet-B0 and CLIP-ViT models perform well in detection and source identification, with CLIP-ViT showing greater robustness against real-world image degradations. \item Our methods achieve competitive or superior results compared to baselines like AEROBLADE and OCC-CLIP, especially in source model identification. \item Data augmentation with perturbations (e.g., Gaussian noise, JPEG compression) significantly improves model generalization and robustness. \end{itemize}

These results highlight the effectiveness of CNN and CLIP encoders in AI-generated content detection. Future work will focus on enhancing feature interpretability for source model attribution.


\bibliography{sample-2col}

\begin{thebibliography}{18}
\expandafter\ifx\csname natexlab\endcsname\relax\def\natexlab#1{#1}\fi
\providecommand{\url}[1]{\texttt{#1}}
\providecommand{\href}[2]{#2}
\providecommand{\path}[1]{#1}
\providecommand{\DOIprefix}{doi:}
\providecommand{\ArXivprefix}{arXiv:}
\providecommand{\URLprefix}{URL: }
\providecommand{\Pubmedprefix}{pmid:}
\providecommand{\doi}[1]{\href{http://dx.doi.org/#1}{\path{#1}}}
\providecommand{\Pubmed}[1]{\href{pmid:#1}{\path{#1}}}
\providecommand{\bibinfo}[2]{#2}
\ifx\xfnm\relax \def\xfnm[#1]{\unskip,\space#1}\fi
\bibitem[{Goodfellow et~al.(2014)Goodfellow, Pouget-Abadie, Mirza, Xu, Warde-Farley, Ozair, Courville, and Bengio}]{gan}
\bibinfo{author}{I.~Goodfellow}, \bibinfo{author}{J.~Pouget-Abadie}, \bibinfo{author}{M.~Mirza}, \bibinfo{author}{B.~Xu}, \bibinfo{author}{D.~Warde-Farley}, \bibinfo{author}{S.~Ozair}, \bibinfo{author}{A.~Courville}, \bibinfo{author}{Y.~Bengio},
\newblock \bibinfo{title}{Generative adversarial nets},
\newblock \bibinfo{journal}{Advances in neural information processing systems} \bibinfo{volume}{27} (\bibinfo{year}{2014}).
\bibitem[{Kingma(2013)}]{auto_encoder}
\bibinfo{author}{D.~P. Kingma},
\newblock \bibinfo{title}{Auto-encoding variational bayes},
\newblock \bibinfo{journal}{arXiv preprint arXiv:1312.6114}  (\bibinfo{year}{2013}).
\bibitem[{Ho et~al.(2020)Ho, Jain, and Abbeel}]{ddpm}
\bibinfo{author}{J.~Ho}, \bibinfo{author}{A.~Jain}, \bibinfo{author}{P.~Abbeel},
\newblock \bibinfo{title}{Denoising diffusion probabilistic models},
\newblock \bibinfo{journal}{Advances in neural information processing systems} \bibinfo{volume}{33} (\bibinfo{year}{2020}) \bibinfo{pages}{6840--6851}.
\bibitem[{OpenAI(2024)}]{dalle}
\bibinfo{author}{OpenAI}, \bibinfo{title}{Dalle-3}, \bibinfo{year}{2024}. \bibinfo{note}{Https://openai.com/index/dall-e-3/}.
\bibitem[{Saharia et~al.(2022)Saharia, Chan, Saxena, Li, Whang, Denton, Ghasemipour, Gontijo~Lopes, Karagol~Ayan, Salimans et~al.}]{imagen}
\bibinfo{author}{C.~Saharia}, \bibinfo{author}{W.~Chan}, \bibinfo{author}{S.~Saxena}, \bibinfo{author}{L.~Li}, \bibinfo{author}{J.~Whang}, \bibinfo{author}{E.~L. Denton}, \bibinfo{author}{K.~Ghasemipour}, \bibinfo{author}{R.~Gontijo~Lopes}, \bibinfo{author}{B.~Karagol~Ayan}, \bibinfo{author}{T.~Salimans}, et~al.,
\newblock \bibinfo{title}{Photorealistic text-to-image diffusion models with deep language understanding},
\newblock \bibinfo{journal}{Advances in neural information processing systems} \bibinfo{volume}{35} (\bibinfo{year}{2022}) \bibinfo{pages}{36479--36494}.
\bibitem[{Rombach et~al.(2022)Rombach, Blattmann, Lorenz, Esser, and Ommer}]{stable_diffusion}
\bibinfo{author}{R.~Rombach}, \bibinfo{author}{A.~Blattmann}, \bibinfo{author}{D.~Lorenz}, \bibinfo{author}{P.~Esser}, \bibinfo{author}{B.~Ommer},
\newblock \bibinfo{title}{High-resolution image synthesis with latent diffusion models},
\newblock in: \bibinfo{booktitle}{Proceedings of the IEEE/CVF Conference on Computer Vision and Pattern Recognition (CVPR)}, \bibinfo{year}{2022}, pp. \bibinfo{pages}{10684--10695}.
\bibitem[{mid(2024)}]{midjourney}
\bibinfo{title}{Midjourney}, \bibinfo{year}{2024}. \bibinfo{note}{Https://www.midjourney.com/}.
\bibitem[{Marra et~al.(2018)Marra, Gragnaniello, Cozzolino, and Verdoliva}]{CNN_detector}
\bibinfo{author}{F.~Marra}, \bibinfo{author}{D.~Gragnaniello}, \bibinfo{author}{D.~Cozzolino}, \bibinfo{author}{L.~Verdoliva},
\newblock \bibinfo{title}{Detection of gan-generated fake images over social networks},
\newblock in: \bibinfo{booktitle}{2018 IEEE conference on multimedia information processing and retrieval (MIPR)}, \bibinfo{organization}{IEEE}, \bibinfo{year}{2018}, pp. \bibinfo{pages}{384--389}.
\bibitem[{Cozzolino et~al.(2024)Cozzolino, Poggi, Corvi, Nie{\ss}ner, and Verdoliva}]{clip_svm}
\bibinfo{author}{D.~Cozzolino}, \bibinfo{author}{G.~Poggi}, \bibinfo{author}{R.~Corvi}, \bibinfo{author}{M.~Nie{\ss}ner}, \bibinfo{author}{L.~Verdoliva},
\newblock \bibinfo{title}{Raising the bar of ai-generated image detection with clip},
\newblock in: \bibinfo{booktitle}{Proceedings of the IEEE/CVF Conference on Computer Vision and Pattern Recognition}, \bibinfo{year}{2024}, pp. \bibinfo{pages}{4356--4366}.
\bibitem[{Radford et~al.(2021)Radford, Kim, Hallacy, Ramesh, Goh, Agarwal, Sastry, Askell, Mishkin, Clark et~al.}]{clip}
\bibinfo{author}{A.~Radford}, \bibinfo{author}{J.~W. Kim}, \bibinfo{author}{C.~Hallacy}, \bibinfo{author}{A.~Ramesh}, \bibinfo{author}{G.~Goh}, \bibinfo{author}{S.~Agarwal}, \bibinfo{author}{G.~Sastry}, \bibinfo{author}{A.~Askell}, \bibinfo{author}{P.~Mishkin}, \bibinfo{author}{J.~Clark}, et~al.,
\newblock \bibinfo{title}{Learning transferable visual models from natural language supervision},
\newblock in: \bibinfo{booktitle}{International conference on machine learning}, \bibinfo{organization}{PMLR}, \bibinfo{year}{2021}, pp. \bibinfo{pages}{8748--8763}.
\bibitem[{Alam et~al.(2024)Alam, Muneer, and Woo}]{fft}
\bibinfo{author}{I.~Alam}, \bibinfo{author}{M.~S. Muneer}, \bibinfo{author}{S.~S. Woo},
\newblock \bibinfo{title}{Ugad: Universal generative ai detector utilizing frequency fingerprints},
\newblock in: \bibinfo{booktitle}{Proceedings of the 33rd ACM International Conference on Information and Knowledge Management}, \bibinfo{year}{2024}, pp. \bibinfo{pages}{4332--4340}.
\bibitem[{Wang et~al.(2023)Wang, Bao, Zhou, Wang, Hu, Chen, and Li}]{dire}
\bibinfo{author}{Z.~Wang}, \bibinfo{author}{J.~Bao}, \bibinfo{author}{W.~Zhou}, \bibinfo{author}{W.~Wang}, \bibinfo{author}{H.~Hu}, \bibinfo{author}{H.~Chen}, \bibinfo{author}{H.~Li},
\newblock \bibinfo{title}{Dire for diffusion-generated image detection},
\newblock in: \bibinfo{booktitle}{Proceedings of the IEEE/CVF International Conference on Computer Vision}, \bibinfo{year}{2023}, pp. \bibinfo{pages}{22445--22455}.
\bibitem[{Ricker et~al.(2024)Ricker, Lukovnikov, and Fischer}]{aeroblade}
\bibinfo{author}{J.~Ricker}, \bibinfo{author}{D.~Lukovnikov}, \bibinfo{author}{A.~Fischer},
\newblock \bibinfo{title}{Aeroblade: Training-free detection of latent diffusion images using autoencoder reconstruction error},
\newblock in: \bibinfo{booktitle}{Proceedings of the IEEE/CVF Conference on Computer Vision and Pattern Recognition (CVPR)}, \bibinfo{year}{2024}, pp. \bibinfo{pages}{9130--9140}.
\bibitem[{Liu et~al.(2024)Liu, Luo, Li, Torr, and Gu}]{occ_clip}
\bibinfo{author}{F.~Liu}, \bibinfo{author}{H.~Luo}, \bibinfo{author}{Y.~Li}, \bibinfo{author}{P.~Torr}, \bibinfo{author}{J.~Gu}, \bibinfo{title}{Which model generated this image? a model-agnostic approach for origin attribution}, \bibinfo{year}{2024}. \href{http://arxiv.org/abs/2404.02697v2}{{\tt arXiv:2404.02697v2}}.
\bibitem[{Tan and Le(2019)}]{efficientnet}
\bibinfo{author}{M.~Tan}, \bibinfo{author}{Q.~Le},
\newblock \bibinfo{title}{Efficientnet: Rethinking model scaling for convolutional neural networks},
\newblock in: \bibinfo{booktitle}{International conference on machine learning}, \bibinfo{organization}{PMLR}, \bibinfo{year}{2019}, pp. \bibinfo{pages}{6105--6114}.
\bibitem[{Cortes(1995)}]{svm}
\bibinfo{author}{C.~Cortes},
\newblock \bibinfo{title}{Support-vector networks},
\newblock \bibinfo{journal}{Machine Learning}  (\bibinfo{year}{1995}).
\bibitem[{Roy et~al.(2025{\natexlab{a}})Roy, Aziz, Bajpai, Imanpour, Singh, Biswas, Wanaskar, Patwa, Ghosh, Dixit, Pal, Rawte, Garimella, Das, Sheth, Sharma, Reganti, Jain, and Chadha}]{roy-2025-defactify-dataset-image}
\bibinfo{author}{R.~Roy}, \bibinfo{author}{A.~Aziz}, \bibinfo{author}{S.~Bajpai}, \bibinfo{author}{N.~Imanpour}, \bibinfo{author}{G.~Singh}, \bibinfo{author}{S.~Biswas}, \bibinfo{author}{K.~Wanaskar}, \bibinfo{author}{P.~Patwa}, \bibinfo{author}{S.~Ghosh}, \bibinfo{author}{S.~Dixit}, \bibinfo{author}{N.~R. Pal}, \bibinfo{author}{V.~Rawte}, \bibinfo{author}{R.~Garimella}, \bibinfo{author}{A.~Das}, \bibinfo{author}{A.~Sheth}, \bibinfo{author}{V.~Sharma}, \bibinfo{author}{A.~N. Reganti}, \bibinfo{author}{V.~Jain}, \bibinfo{author}{A.~Chadha},
\newblock \bibinfo{title}{Defactify-image: A comprehensive dataset for human vs. ai generated image detection},
\newblock in: \bibinfo{booktitle}{proceedings of {D}e{F}actify 4: Fourth workshop on Multimodal Fact-Checking and Hate Speech Detection}, \bibinfo{publisher}{CEUR}, \bibinfo{year}{2025}{\natexlab{a}}.
\bibitem[{Roy et~al.(2025{\natexlab{b}})Roy, Imanpour, Aziz, Bajpai, Singh, Biswas, Wanaskar, Patwa, Ghosh, Dixit, Pal, Rawte, Garimella, Das, Sheth, Sharma, Reganti, Jain, and Chadha}]{roy-2025-defactify-overview-image}
\bibinfo{author}{R.~Roy}, \bibinfo{author}{N.~Imanpour}, \bibinfo{author}{A.~Aziz}, \bibinfo{author}{S.~Bajpai}, \bibinfo{author}{G.~Singh}, \bibinfo{author}{S.~Biswas}, \bibinfo{author}{K.~Wanaskar}, \bibinfo{author}{P.~Patwa}, \bibinfo{author}{S.~Ghosh}, \bibinfo{author}{S.~Dixit}, \bibinfo{author}{N.~R. Pal}, \bibinfo{author}{V.~Rawte}, \bibinfo{author}{R.~Garimella}, \bibinfo{author}{A.~Das}, \bibinfo{author}{A.~Sheth}, \bibinfo{author}{V.~Sharma}, \bibinfo{author}{A.~N. Reganti}, \bibinfo{author}{V.~Jain}, \bibinfo{author}{A.~Chadha},
\newblock \bibinfo{title}{Overview of image counter turing test: Ai generated image detection},
\newblock in: \bibinfo{booktitle}{proceedings of {D}e{F}actify 4: Fourth workshop on Multimodal Fact-Checking and Hate Speech Detection}, \bibinfo{publisher}{CEUR}, \bibinfo{year}{2025}{\natexlab{b}}.

\end{thebibliography}

\appendix

\end{document}